\documentclass[conference]{IEEEtran}
\IEEEoverridecommandlockouts
\usepackage{cite}
\usepackage{amsmath,amssymb,amsfonts}
\usepackage{algorithmic}
\usepackage{graphicx}
\usepackage{textcomp}
\usepackage{xcolor}
\def\BibTeX{{\rm B\kern-.05em{\sc i\kern-.025em b}\kern-.08em
    T\kern-.1667em\lower.7ex\hbox{E}\kern-.125emX}}

\usepackage{url}
\usepackage{tablefootnote}
\usepackage{float}
\newcommand{\quotes}[1]{``#1''}

\begin{document}

\title{Towards A Foundation Model For Trajectory Intelligence}

\author{\IEEEauthorblockN{Alameen Najjar}
\IEEEauthorblockA{\textit{Rakuten Institute of Technology} \\
Tokyo, Japan \\
alameen.najjar@rakuten.com}
}

\maketitle

\begin{abstract}
We present the results of training a large trajectory model using real-world user check-in data. Our approach follows a pre-train and fine-tune paradigm, where a base model is pre-trained via masked trajectory modeling and then adapted through fine-tuning for various downstream tasks. To address challenges posed by noisy data and large spatial vocabularies, we propose a novel spatial tokenization block. Our empirical analysis utilizes a comprehensive dataset of over 2 billion check-ins generated by more than 6 million users. Through fine-tuning on 3 downstream tasks we demonstrate that our base model has effectively learned valuable underlying patterns in raw data, enabling its application in meaningful trajectory intelligence tasks. Despite some limitations, we believe this work represents an important step forward in the realization of a foundation model for trajectory intelligence.
\end{abstract}


\section{Introduction}
\label{sec1}

Recently, there has been a surge of interest in trajectory foundation models, also referred to as large trajectory models~(LTMs), within the geospatial artificial intelligence community \cite{musleh2022let, musleh2022towards, mai2022towards, musleh2023demonstration}. These specialized models undergo extensive task-agnostic training on trajectory data, granting them a profound understanding of how humans interact with the physical space. Consequently, this broad domain knowledge can be fine-tuned for various trajectory intelligence tasks. Despite the growing interest, the concrete realization of an actual LTM is still awaited \cite{musleh2022let}.

Although trajectory data shares inherent similarities with natural language \cite{musleh2022let}, training LTMs using techniques similar to those used for large language models (LLMs) poses several challenges. One major obstacle is the limited availability of trajectory data; the largest publicly accessible trajectory dataset~\cite{yang2016participatory} contains only a few tens of millions of data points, a stark contrast to the trillions of tokens used to train the latest LLMs~\cite{anil2023palm, touvron2023llama}. Handling inflated spatial vocabularies is another major technical challenge (See Section \ref{sec2}).

\begin{figure}[ht]
  \centering
  \includegraphics[width=1\linewidth]{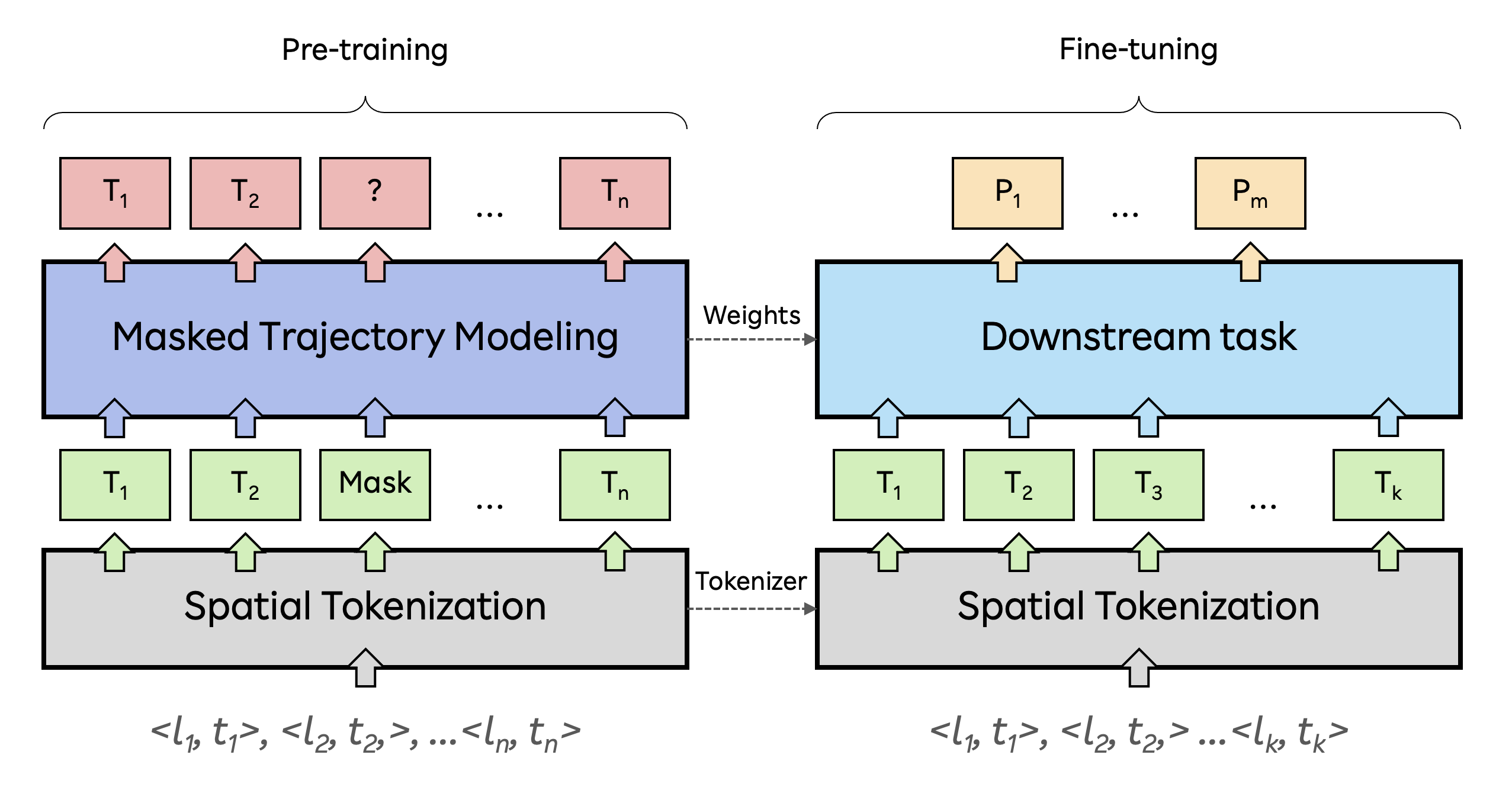}
  \caption{The adopted pre-train and fine-tune paradigm. Masked trajectory modeling is utilized to learn a base model which is later fine-tuned for a variety of trajectory intelligence tasks.}
  \label{fig1}
\end{figure}

In this work, we attempt to train a large trajectory model using real-world check-in data. We adopt a pre-train and fine-tune paradigm (See Figure \ref{fig1}), where a base model is pre-trained via masked trajectory modeling and then adapted through fine-tuning for various downstream tasks. Utilizing a comprehensive dataset of over 2 billion check-ins generated by more than 6 million users, and through fine-tuning, we demonstrate that our base model has effectively learned valuable underlying patterns in raw data, enabling its application in meaningful trajectory intelligence tasks.

The contributions we make in this work can be outlined as follows:
\begin{enumerate}
    \item Training an LTM on over 40 billion spatial tokens, which, to the best of our knowledge, has never been attempted before.
    \item Proposing a novel spatial tokenization block that effectively handles noisy trajectory data and addresses the difficulties arising from large spatial vocabularies (See Section \ref{sec2} for details).
\end{enumerate}

The remainder of this manuscript is structured as follows: In Section \ref{sec2}, we elaborate on our approach to LTMs. Experiments are detailed in Section \ref{sec3}. Finally, Section \ref{sec4} presents a summary of our work and outlines potential directions for future research.

\section{Large Trajectory Model}
\label{sec2}

\subsection{Pre-train and Fine-tune Paradigm}

Our approach to LTMs follows the pre-train and fine-tune paradigm, a common strategy for building LLMs. The process involves two distinct stages: pre-training and fine-tuning (See Figure \ref{fig1}).

In the pre-training stage, a base model is trained on an extensive corpus of raw trajectory data using unsupervised learning. The model is trained in a task-agnostic manner, enabling it to gain a broad understanding of trajectory data as a modality.

Subsequently, during the fine-tuning stage, the model is specialized for specific trajectory intelligence tasks by training it on labeled data from the target task. The model's parameters are adjusted to optimize its performance on the target task, allowing it to adapt its knowledge to the specific domain or problem at hand.

This paradigm proves advantageous as it harnesses the abundance of data available during pre-training to capture general patterns and nuances of human mobility. Fine-tuning then tailors this knowledge to specific tasks, resulting in an adaptable foundation model capable of achieving excellent performance across various trajectory intelligence tasks.

\subsection{Masked Trajectory Modeling}

The pre-training of our LTM involves the application of Masked Trajectory modeling (MTM) as a key task. MTM entails the masking of a random subset of points within a trajectory, prompting the model to predict the missing points based on the bidirectional contextual information. This methodology draws parallels with Masked Language Modeling~(MLM) in the realm of natural language processing.

By employing MTM, our objective is to enable the model to develop a comprehensive understanding of how people interact with the physical space. This understanding is derived from the underlying patterns present in the raw data.

\subsection{Spatial Tokenization}

\begin{figure}[ht]
  \centering
  \includegraphics[width=1\linewidth]{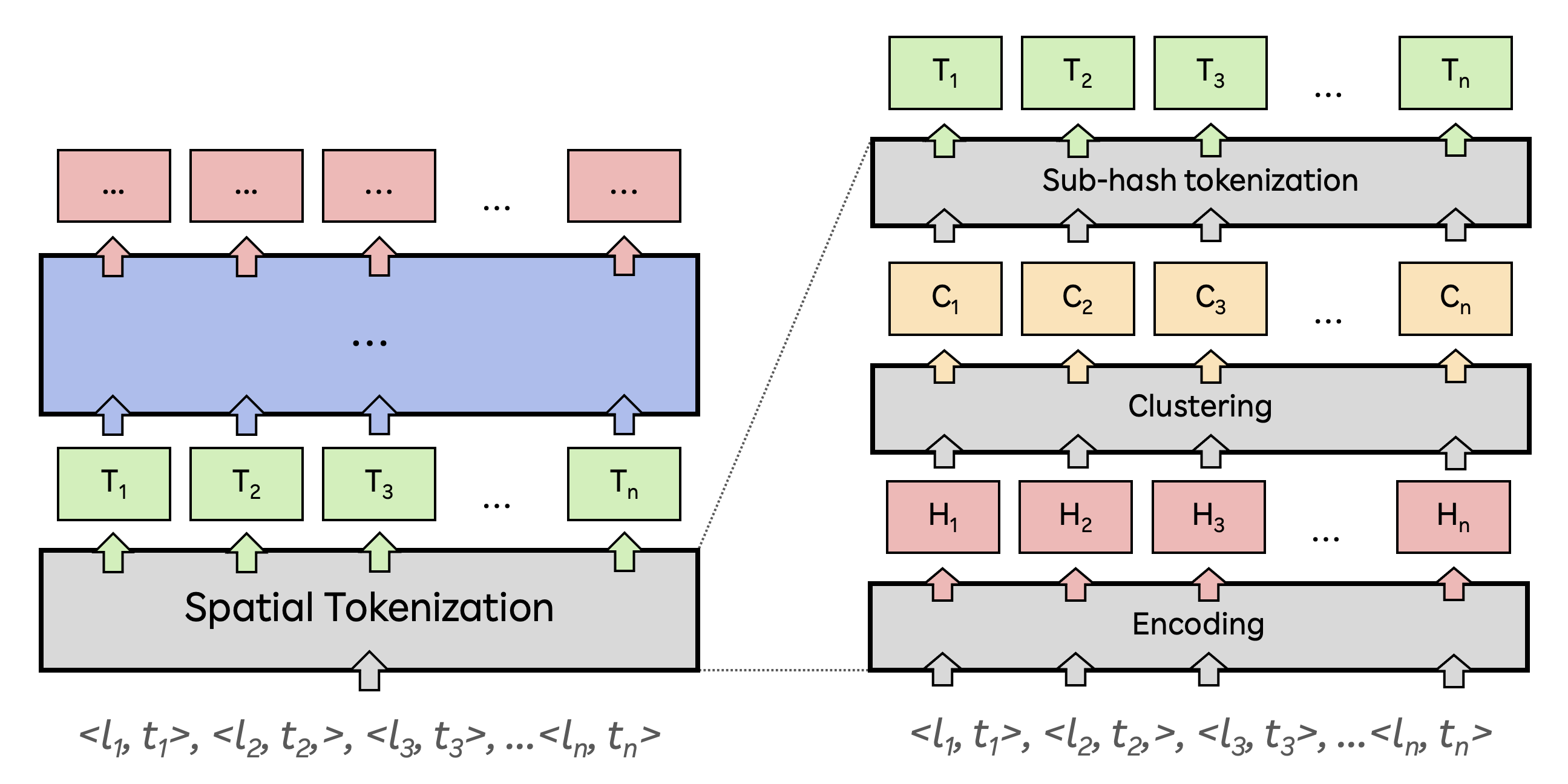}
  \caption{Under the hood look at the proposed spatial tokenizer. Raw trajectory data is transformed into spatial tokens in three consecutive steps: 1) encoding, 2) clustering, and~3)~sub-hash tokenization.}
  \label{fig2}
\end{figure}

We propose a novel spatial tokenizer (Figure \ref{fig2}) designed to effectively handle noisy trajectory data and address challenges associated with large spatial vocabularies. The tokenizer's operation involves three consecutive steps: \bigbreak

\noindent\emph{Encoding}. Raw trajectory data, comprising latitude and longitude pairs, is discretized through a mapping process into a hash representation that corresponds to a distinct earth surface region. This can be achieved using off-the-shelf solutions such as Uber H3 \cite{brodsky2018h3}, Google S2\footnote{\url{https://s2geometry.io/}}, or Geohash\footnote{\url{http://geohash.org/}}. \bigbreak

\noindent\emph{Clustering}. The hash sequences produced in the previous step, which also include timestamps, are organized into clusters based on spatiotemporal proximity. This step serves to control the density of trajectory points, with the goal of minimizing noise. \bigbreak

\begin{figure}[ht]
  \centering
  \includegraphics[width=1\linewidth]{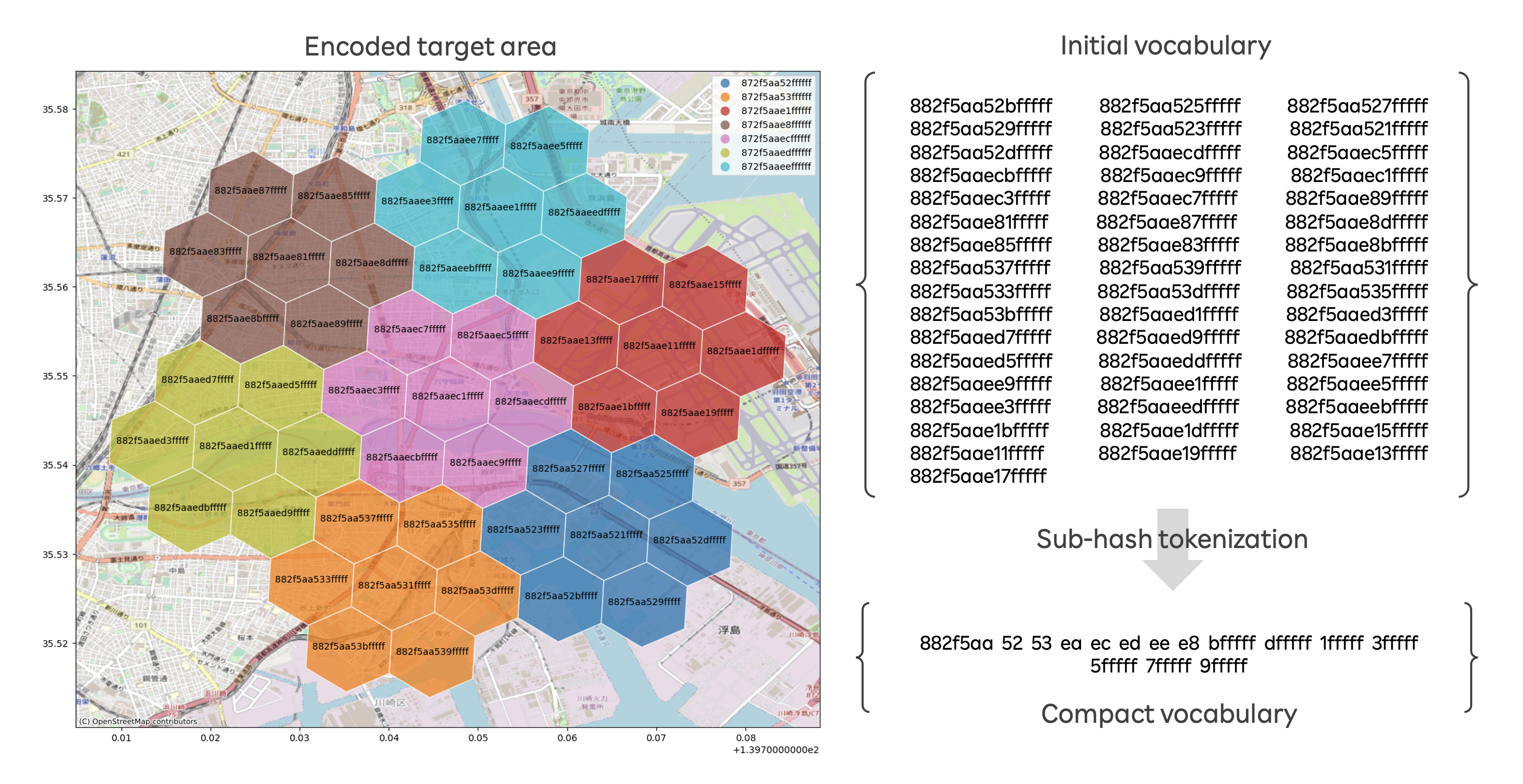}
  \caption{Illustrative example of the proposed sub-hash tokenization. Thanks to hierarchical hashing, the initial vocabulary of 49 unique hashes is reduced down to a vocabulary of 15 sub-hashes only. \emph{It is worth noting that it takes over 500,000 unique hashes to encode a country the size of Japan. Which is a vocabulary significantly larger than that of the largest LLMs} \cite{xue2020mt5}.}
  \label{fig3}
\end{figure}

\noindent\emph{Sub-hash tokenization}. Leveraging their hierarchical nature, clustered hashes obtained earlier are divided into sub-hashes. \emph{This step is crucial because it helps manage the inflated spatial vocabulary linked to the increasing geographic area, and consequently reduces the computational requirements for both training and inference}. See Figure \ref{fig3} for an illustrative example. \bigbreak

\begin{figure}[H]
  \centering
  \includegraphics[width=1\linewidth]{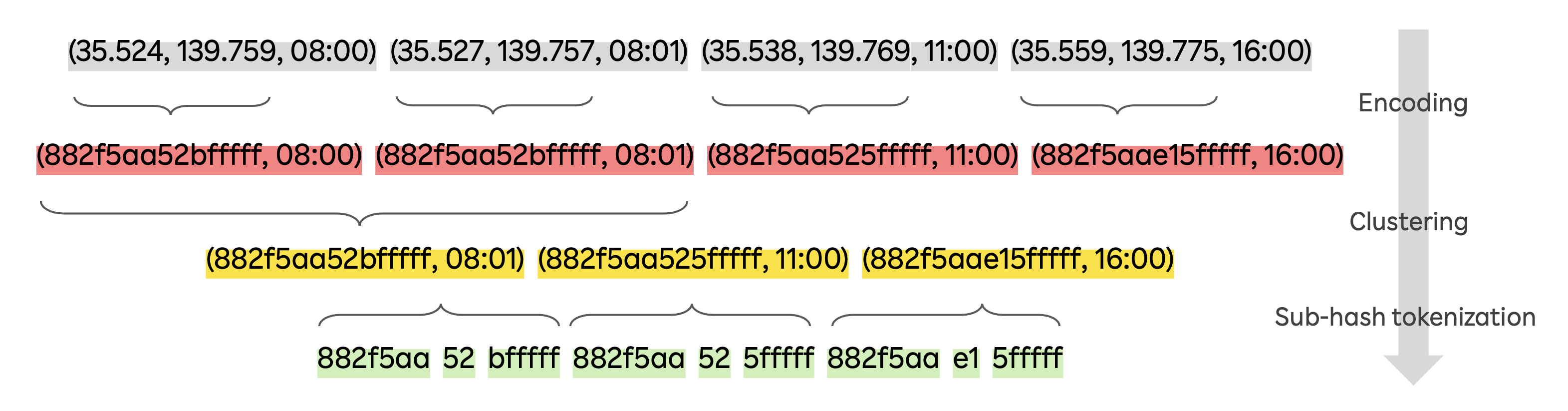}
  \caption{Illustrative example of spatial tokenization applied to an 8-hour (08:00-16:00) trajectory. Raw trajectory, hashes, clusters and tokens are highlighted in gray, red, yellow and green, respectively.}
  \label{fig4}
\end{figure}

Overall, the proposed spatial tokenizer efficiently processes raw trajectory data by transforming it into spatial tokens. This process involves encoding, clustering, and sub-hash tokenization. See Figure~\ref{fig4} for an illustrative example.

\section{Experiments}
\label{sec3}

\subsection{Data}
\begin{figure}[ht]
  \centering
  \includegraphics[width=0.8\linewidth]{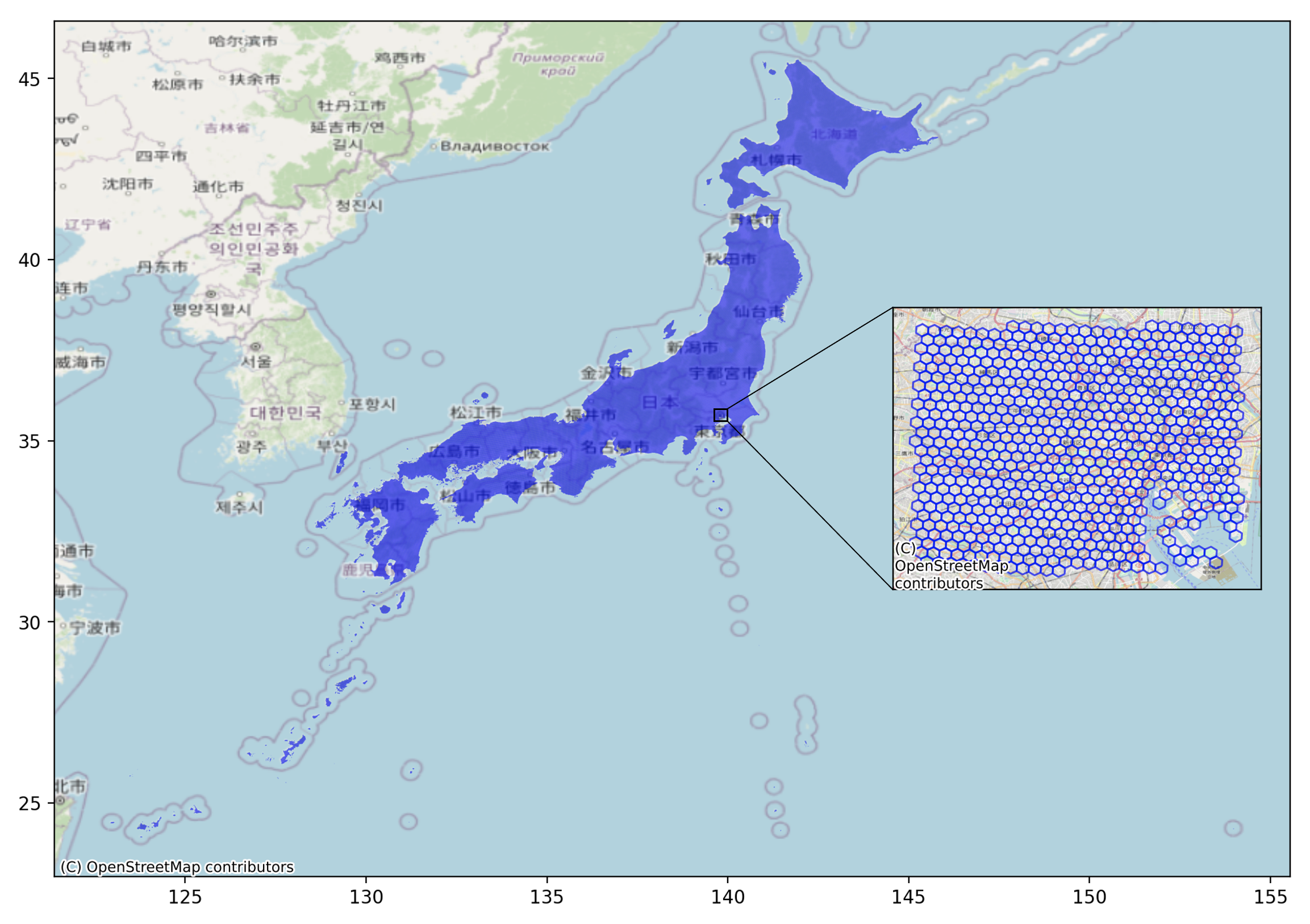}
  \caption{Polygons generated by aggregating individual check-ins naturally trace out the the political boundary of Japan.}
  \label{fig5}
\end{figure}

For all subsequent analysis we utilize a proprietary check-in dataset generated by over 6 million users of different Rakuten\footnote{\url{https://www.rakuten.co.jp/}} services active in Japan over a span of 12 months. Each check-in is represented as an anonymized user ID, latitude/longitude pair, and timestamp tuple. Overall, the dataset comprises over~2 billion check-ins, notably surpassing the size of any publicly accessible dataset of its kind that we are aware of. See Figure \ref{fig5} for a map of the dataset's coverage, and Table \ref{tab1} for a comparison with existing datasets.

\begin{table}
\centering
  \caption{Dataset comparison: Although limited to Japan, our dataset is significantly larger than all publicly accessible check-in datasets combined.}
  \label{tab1}
  \begin{tabular}{|l|lll|}
    \hline
    Dataset     & Check-ins & Users & Date range \\
    \hline
    Brightkite \cite{cho2011friendship}  & 4.7M      & 51.4K & 3/2008 - 10/2010 \\
    Gowalla \cite{cho2011friendship}     & 6.4M      & 107K  & 2/2009 - 10/2010 \\
    Weeplaces\tablefootnote{\url{https://www.yongliu.org/datasets/}}  & 7.4M      & 15.7K & 11/2003 - 6/2011 \\
    Foursquare \cite{yang2016participatory}  & 33.2M     & 267K  & 4/2012 - 9/2013 \\
    Ours        & 2B+     & 6M+  & 7/2022 - 6/2023 \\
  \hline
\end{tabular}
\end{table}

\subsection{Implementation}
\emph{Data Processing}. We started out by organizing user check-ins into monthly trajectories. Trajectories spanning fewer than~3 distinct hashes and those with a length below~10 are excluded. Next, user IDs are eliminated for privacy. These trajectories are then divided into three sets: 70\% for training, 15\% for validation, and 15\% for testing purposes. \bigbreak

\noindent\emph{Spatial Tokenizer}. We employed Uber H3 (Resolution 8) as the location encoder. We set the spatiotemporal constraints for clustering to 10 minutes and one H3 hexagon. Finally, we trained a sub-hash tokenizer with a 30k vocabulary using the WordPiece algorithm \cite{wu2016google}. \bigbreak

\noindent\emph{Transformer}. We employed the Hugging Face \cite{wolf2019huggingface} implementation of BERT \cite{devlin2018bert} as the encoder, utilizing the default hyper-parameter settings unless otherwise noted. \bigbreak

\noindent\emph{Pre-training}. We applied whole-hexagon pre-masking to both the training and validation data splits, using a 20\% masking ratio. Pre-training ran for 40 epochs, spanning approximately 240 hours, and was executed on 2 NVIDIA Tesla V100 GPUs running in parallel. Chunk size and batch size were set at 512 and 64, respectively. \bigbreak

\noindent\emph{Fine-tuning}. Only the test split is used in fine-tuning. A random~20k examples are chosen per downstream task which is further split into 80\% and 20\% balanced subsets for training and validation, respectively. We ran fine-tuning for 10 epochs only using a batch size of 64.

\begin{figure}[ht]
  \centering
  \includegraphics[width=0.95\linewidth]{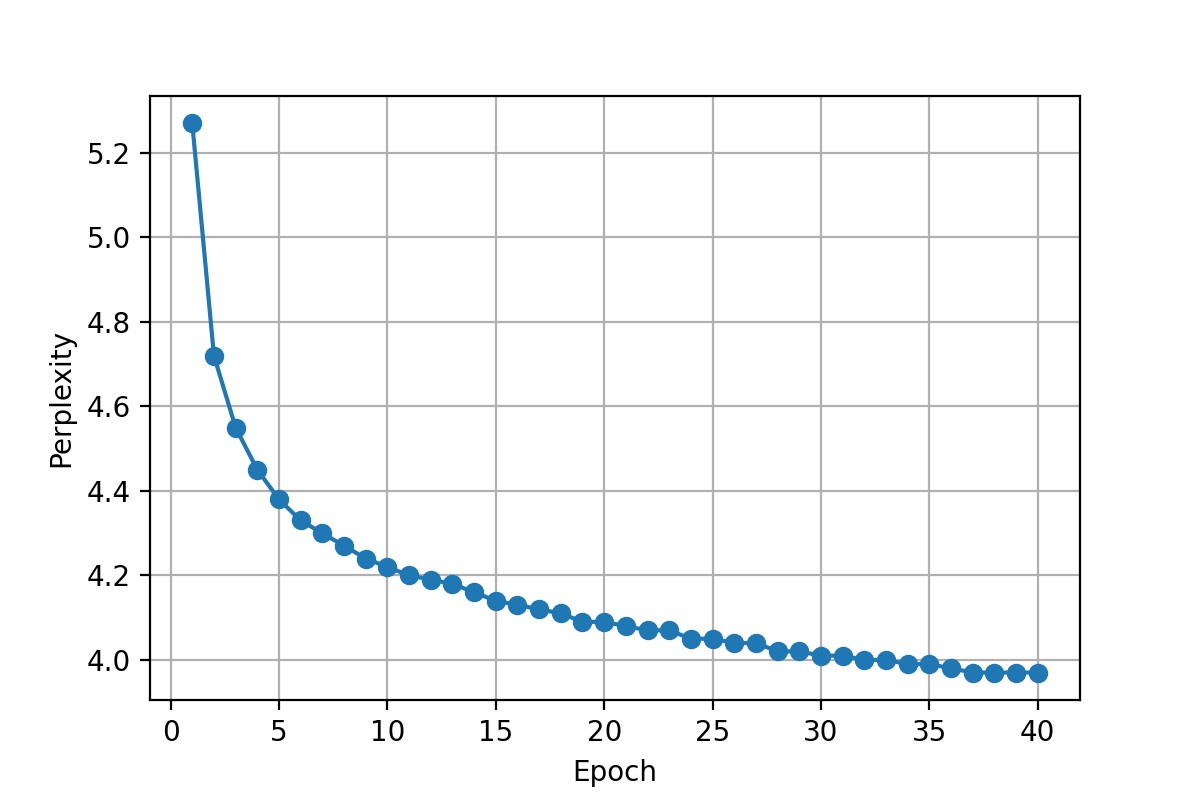}
  \caption{Pre-training evaluation: Validation perplexity plotted over time.}
  \label{fig6}
\end{figure}

\subsection{Results}

To evaluate the quality of the pre-training, in Figure \ref{fig6}, we plotted the model's validation perplexity over time. Perplexity measures how \quotes{surprised} the model is when it sees new data, and it is a commonly used metric in natural language processing \cite{rabiner1989tutorial}. After 40 epochs of training, perplexity reaches~3.79 which is a significant drop from the initial (Random initialization) value of 2327. It is worth noting that by the end of the first epoch the model has already seen over a billion spatial tokens, which might explain the initial relatively low perplexity value of 5.27. We further investigated this theory by pre-training a new model using a significantly smaller dataset of roughly 80 million tokens and we found that the perplexity gradually drops to a comparable value ($\approx8$) by the end of the~12th epoch. Which supports our hunch regarding the initially relatively low perplexity value.

While this experiment demonstrates that our model has gotten better at predicting missing points in trajectory data, it does not indicate whether the model has managed to grasp valuable underlying patterns in trajectory data. To that end, we need to evaluate the model's generalizability on relevant downstream tasks, which we turn to next.

\begin{figure}[ht]
  \flushleft
  \includegraphics[width=1\linewidth]{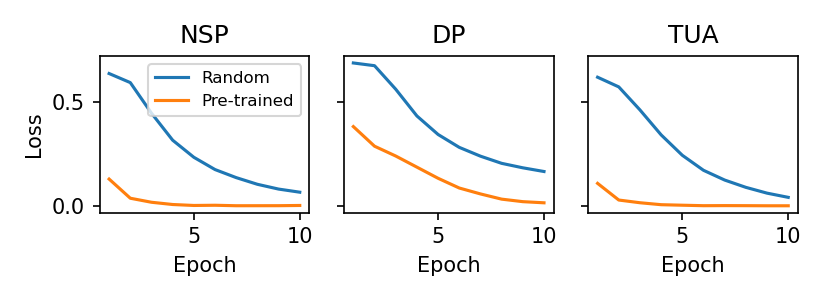}
  \caption{Fine-tuning evaluation (2): Training loss curves of three different downstream tasks. \quotes{NSP,} \quotes{DP,} and \quotes{TUA} stand for \quotes{next sub-trajectory prediction,} \quotes{destination prediction} and \quotes{trajectory-user association}, respectively.}
  \label{fig7}
\end{figure}

In Table \ref{tab2} and Figure \ref{fig7} we evaluate the model's fine-tuning performance on three downstream tasks, namely next sub-trajectory prediction (NSP), destination prediction (DP) and Trajectory-User association (TUA). In NSP, the model is asked to predict whether a sub-trajectory follows the one before or not. While in DP, the model is asked to predict the destination~(Last visited hexagon in a trajectory) given the first~25\% of the trajectory. Finally, in TUA, the model is asked whether two trajectories are generated by the same user or not. On all three tasks, and after 10 epochs of fine-tuning, the pre-trained model significantly outperforms (33.6\% better) the fine-tuned model initialized with random weights. 

These results demonstrate that the model has captured valuable underlying patterns in the raw data which enabled it to generalize well with fine-tuning.

\begin{table}
\centering
  \caption{Fine-tuning evaluation (1): Average F1-score obtained after 10 epochs of fine-tuning. \quotes{Random} and \quotes{Pre-trained} indicate randomly assigned and pre-trained weights, respectively.}
  \label{tab2}
  \begin{tabular}{|l|ll|}
    \hline
    Task                                 & Random & Pre-trained \\
    \hline
    Next sub-trajectory prediction       & 0.62   & 0.94 \\
    Destination prediction               & 0.55   & 0.84 \\
    Trajectory-user association          & 0.57   & 0.97 \\

  \hline
\end{tabular}
\end{table}

\section{Summary and Future Work}
\label{sec4}

In this paper, we presented the results of training a large trajectory model on real-world user check-in data. We adopt a pre-train and fine-tune paradigm, where a base model is pre-trained via masked trajectory modeling and then adapted through fine-tuning for various downstream tasks. Utilizing a comprehensive dataset of over 2 billion check-ins, and through fine-tuning, we demonstrated that our base model has effectively learned valuable underlying patterns in raw data, enabling its application in meaningful trajectory intelligence tasks.

One of our key contributions is the introduction of a novel spatial tokenizer block, comprising encoding, spatiotemporal clustering, and sub-hash tokenization. This innovative tokenizer allows us to treat trajectory data like natural language, effectively addressing noisy data and the challenge of large spatial vocabularies. Furthermore, our experiments utilize a substantially larger dataset compared to any publicly available check-in dataset, enhancing the robustness and generalizability of our findings.

Acknowledging the ethical implications of developing such a large trajectory model, we emphasize the sensitivity of location data, as owning and deploying such a model based on the offline behavior of millions of users can lead to potential privacy and ethical concerns. To mitigate these risks, we took great care in preserving user privacy by anonymizing user IDs and aggregating locations using a mid-resolution grid.

Despite the success of our approach, there are opportunities for further advancements. A primary avenue for future work is incorporating time into our trajectory model. Currently, the model is spatially aware but does not explicitly consider the temporal dimension, which is essential for understanding dynamic user behaviors and the evolution of trajectories over time. By incorporating time into the model, we can enhance the predictive capabilities and gain insights into how users' movements change over different periods.

Additionally, exploring ways to probe the underlying patterns in the raw data learned by the model during pre-training remains a critical area for future research.

In conclusion, despite some inherent limitations, we believe this work represents an important step forward in the realization of a foundation model for trajectory intelligence. By addressing the spatial and temporal aspects of user movements and being attentive to ethical considerations, future research can continue to advance trajectory modeling, benefiting a wide range of applications, including urban planning, transportation, and location-based services.

\section*{Acknowledgment}
The author would like to sincerely thank Kyle Mede for his insightful feedback on the final version of the manuscript.

\bibliographystyle{IEEEtran}
\bibliography{bibs}

\end{document}